# AUTOMATIC TEXT EXTRACTION AND CHARACTER SEGMENTATION USING MAXIMALLY STABLE EXTREMAL REGIONS

Nitigya Sambyal
Department of Computer Science & I.T
University of Jammu, Jammu,
Jammu and Kashmir, India, 180006
E-mail: sambyal.nitigya@gmail.com

Pawanesh Abrol*
Department of Computer Science & I.T
University of Jammu, Jammu,
Jammu and Kashmir, India, 180006
E-mail: pawanesh.abrol@gmail.com
*corresponding author

**Abstract:** Text detection and segmentation is an important prerequisite for many content based image analysis tasks. The paper proposes a novel text extraction and character segmentation algorithm using Maximally Stable Extremal Regions as basic letter candidates. These regions are then subjected to thresholding and thereafter various connected components are determined to identify separate characters. The algorithm is tested along a set of various JPEG, PNG and BMP images over four different character sets; English, Russian, Hindi and Urdu. The algorithm gives good results for English and Russian character set; however character segmentation in Urdu and Hindi language is not much accurate. The algorithm is simple, efficient, involves no overhead as required in training and gives good results for even low quality images. The paper also proposes various challenges in text extraction and segmentation for multilingual inputs.

## I. INTRODUCTION

Text extraction and segmentation is an important phase in document recognition system and is widely used form of human beings language. As more and more information is transformed in digital form, visual texts are embedded in many forms of digital media such as images, documents, videos etc. This text present important information like names, locations, brands of products etc. which facilitate ease in indexing and understanding. The text embedded in media is of two types: Superimposed text and scene text. Superimposed text are also known as graphical text, is horizontally aligned and present in foreground with size, font spacing and orientation remaining constant within a text region. Whereas scene text is present in recorded scene or photograph with variable size, font, color, alignment, lightening condition etc. [1]

Text detection and recognition can play vital role in day to day life of humans and in future could be part of so many computer applications. More over text extraction plays a crucial role for blind people when they want to read the text presented in the scene images [9]. It will enable in indexing the documents having high level of visual content by using text based search engines [1][13].It can help in developing systems to count characters, words in a document, identify the type of script, thereby helping in paleography study. It can help in distinguishing between images and documents with embedded visual content and developing text conversion system, thus opening the possibility for more improved, efficient, fast and advanced systems [14].

Text detection, which is identification of textual regions and segmentation system which is segregating, each characters in the textual region is designed to elicit different types of text from arbitrary complex images, documents or videos. To segment various characters in the document, it is imperative first to identify all the manuscript regions in the document. Identifying text is in itself a complex process as text characteristics can vary in font, size, style, orientation, alignment, contrast, texture, color and background information. Correctness of text line segmentation directly affects accuracy of segmentation of word which subsequently affects the accuracy of character recognition. There are various techniques for text extraction and segmentation in images e.g. text segmentation based on texture aims at finding the difference between the text pixels and the background pixels. Text segmentation can be done by training the neural network following which it is used to classify grey level pixels into text pixels and non- text pixels. Text strings can be further be refined by applying certain constraints like edge detectors, region filtering, stroke width etc. [4].

Although many research efforts have been made to detect text regions in the images the paper proposes a novel text detection and segmentation algorithm which employs Maximally Stable Extremal Regions as basic letter candidates. MSER helps in identifying character regions even when the image is of low quality [8]. Following this the characters are segmented using connected component approach.

The paper is organized as follows: section II briefly overviews the various text detection and segmentation algorithms. Section III describes the proposed approach for extracting text and character candidates. Section IV presents the results and analysis of the algorithm on various test samples. Section V consists of conclusion along with various challenges.

## II. RELATED WORK

Text detection and recognition involves searching of all characters in the document. This can be thought of sliding a window across the document and finding the characters in it. A boosting framework for object detection was proposed by Viola and Jones [3]. Their framework was based on Adaboost providing high accuracy and speed. This learning framework was later extended to work with other classes of objects such as license plates. Text line segmentation can be done by morphology and histogram projection. In this process, firstly a Y histogram projection is performed which results in the text lines positions. Threshold is applied to divide the lines in different regions.

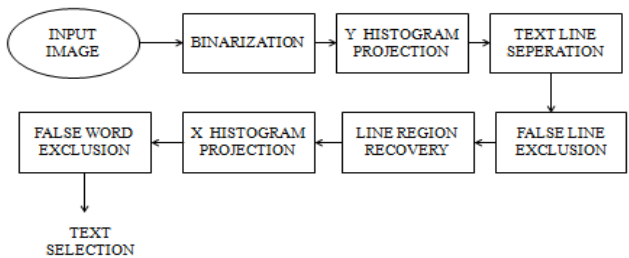

Figure I: Steps in text line segmentation based on morphology & histogram projection

After this, another threshold is used to eliminate false lines. These procedures, however, causes some loss on the text line area. To overcome such losses, a recovery method is proposed to minimize this effect. In order to detect the extreme positions of the text in the horizontal direction, an X histogram projection is applied. Another threshold in the Y direction is used to eliminate false words. Finally, in order to optimize the area of the manuscript text line, a text selection is carried out [5]. The various steps involved in the process can be represented as shown in Fig I.

Bounded box method can be used for segmenting lines, words and characters in a document. It is based on the pixel histogram approach where text images are converted to binarised images by thresholding using Otsu's method [10] and then numbers of text lines are detected using horizontal histogram. The number of peaks in the histogram corresponds to number of text lines. This is followed by plotting vertical histogram which enables segmenting words in each text line. Finally, each character in the word is segmented after applying thinning operation on the image [6].

A segmentation algorithm is proposed by Ohya wherein documents are segmented into regions and character, candidate regions are selected through checking features under assumptions like width, gray level of text etc. Recognition process is then applied to cluster the separate parts of one text character together and extract character pattern candidates [1].

Another algorithm is used to segment documents into text regions and background. Text regions are refined under the constraints such as height similarity, spacing and alignment. Thereafter, these are binarised and the constraints are further reapplied with tighter standards [1].

A technique for segmentation with neural network is also proposed, where a neural network is trained with the output of multiscale wavelets extracted from the regions belonging to the foreground and background regions. A machine learning program is trained to discriminate three main categories: Halftone, background and text & line drawing regions. A three layer neural network is used to classify grey level pixels into text pixels and non-text pixels [1]. The system thus, developed can be used for text based image indexing. Such text also serves as important source of content oriented processing. Detection of text string can also be done on local gradient features and color uniformity of character components to find the character candidates [11].

Detecting text in natural images can be done by employing edge-enhanced Maximally Stable Extremal Regions as basic letter candidates. These candidates are then filtered using geometric and stroke width information to exclude non-text objects. Thereafter characters are paired to identify text lines, which are subsequently separated into words [12].

In comparison to the various text detection and segmentation approaches discussed above, the proposed algorithm offers following advantages. First, since it is based on the Maximally Stable Extremal Regions method, it enables extracting a comprehensive number of corresponding image elements which contributes to the wide-baseline matching, and has led to better stereo matching and object recognition algorithms. Second, it can be used to extract feature descriptors for visual search. Hence our text detection can be combined with visual search systems without further computational load to detect interest regions. Finally the proposed algorithm is simple, efficient and involves no overhead as required in training.

## III. PROPOSED FRAMEWORK

The text detection system serves as basis for developing various advanced systems but due to the large number of possible variations in text in different types of applications and the limitations of any single approach to deal with such variations, a hybrid approach is proposed. In this hybrid scheme, the connected component based method is applied after the bounding boxes have been localized as Maximally Stable Extremal Regions (MSER). For testing the accuracy of

The proposed approach multilingual inputs in JPEG, PNG and BMP formats are selected. The various steps involved in

the proposed system are as follows:
1. Load the image.
2. Detect the MSER regions using MSER Feature detector.
3. Use sobel edge detector to further segment the text.
4. Filter character candidates using the connected component analysis.
5. Determining bounding boxes enclosing text regions.
6. For each bounding box obtained, make a label matrix that contains labels for the 8 connected objects found in it.
7. Ascertain the number of connected components using label matrix.
8. Determine the bounding boxes for each of the connected object, thereby segmenting the characters of text lines.

In this approach firstly the colored input images are converted to gray image to eliminate hue and saturation information but retaining the luminance information. The gray image thus obtained, is given as input to Maximally Stable Extremal Region Feature detector which detects the Maximally Stable Extremal Regions. Thereafter sobel edge detector is used to enhance the edges of MSER in the images to distinguish them from other regions. These text regions are enclosed by bounding boxes, as shown in Fig II. Each bounding box is then given as input to character segmentation process. For each bounding box obtained the character segmentation algorithm makes a label matrix that contains labels for the 8 connected objects. Also the total number of components in each bounding box is determined. After the various properties of each connected component is measured, the smallest rectangle containing a single connected component is obtained. The output of character segmentation algorithm is shown in Figure III.

## IV. RESULTS AND DISCUSSIONS

The paper aims at developing a hybrid approach for text detection and character segmentation in images using MSER as basic letter candidates and connected component method for character segmentation. For this purpose, a set of images consisting of JPEG, PNG and BMP in four different character sets: English, Russian, Urdu and Hindi were collected and tested to find the accuracy of the proposed approach. These four multilingual inputs are selected as they vary in shapes, formation and writing style. Table 1 shows the output of text extraction and segmentation for English character set using MSER and connected component method. It is observed that text detection and character segmentation is done correctly both with and without background color for all test cases of English character set

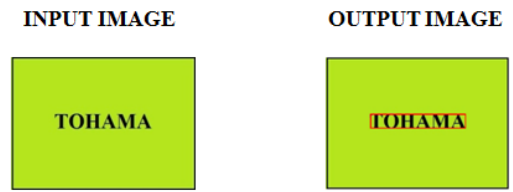

Figure II: Implementation of text detection algorithm using MSER Feature detector

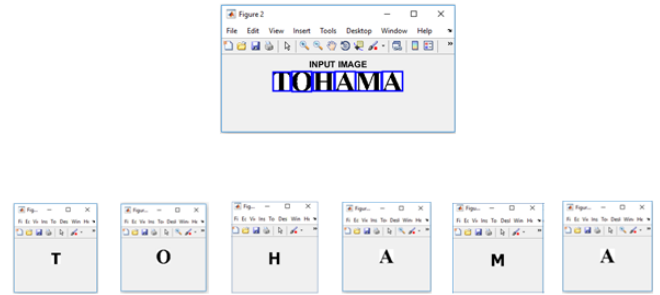

Figure III: Implementation of character segmentation algorithm

Table 2 shows the output of text extraction and segmentation for Russian character set using MSER and connected component method. It is observed that text detection and character segmentation is done correctly with plain white background and colored background. Table 3 shows the output of text extraction and segmentation for Urdu character set using MSER and connected component method. The text detection algorithm shows fairly good accuracy but sometimes the nuqta (.) are mistaken as noise and are not detected as in the test case T13 in Table III.

In Character segmentation the disconnected components of a single character are identified as different characters .This happens as the character segmentation proceeds by identifying connected components as a single character in the given textual region. This is evident from the test case T12 of Table III. Table IV shows the output of text extraction and segmentation for Hindi character set using MSER and connected component method. The text detection algorithm shows fairly good accuracy with and without background color but sometimes the anuswara (.) are mistaken as noise and are not detected as in the test case T18. However character segmentation for Hindi character set reduces to word segmentation due to presence of overline ('Shirorekha')which connects each character underneath it giving a single component. The half letters which are not connected to overline are identified as separate characters as shown in test case T18 in Table IV. Thus it is observed that text detection gives fairly good results for superimposed text in the images with and without background color for the four character sets under study; however character segmentation doesn't provide satisfiable results for Hindi and Urdu Language.

**TABLE I: EXTRACTION OF ENGLISH CHARACTER SET BASED ON MSER AND CONNECTED COMPONENT METHOD**

| Test case number | Input | Output of text detection | Output of character segmentation |
|---|---|---|---|
| T1 | MAT | MAT | M A T |
| T2 | HOUSE | HOUSE | H O U S E |
| T3 | NEW CAR | NEW CAR | N E W<br>C A R |
| T4 | BIRD | BIRD | B I R D |
| T5 | A BOY | A BOY | A<br>B O Y |

**TABLE II: EXTRACTION OF RUSSIAN CHARACTER SET BASED ON MSER AND CONNECTED COMPONENT METHOD**

| Test case number | INPUT | OUTPUT OF TEXT DETECTION | OUTPUT OF CHARACTER SEGMENTATION |
|---|---|---|---|
| T6 | САМ | САМ | С А М |
| T7 | ДВА ДРУГА | ДВА ДРУГА | Д В А<br>Д Р У Г А |
| T8 | сложный | сложный | С Л О Ж Н<br>Ы І И ⌣ |
| T9 | ДОМ | ДОМ | Д О М |
| T10 | ТАК ЛЕГКО | ТАК ЛЕГКО | Т А К<br>Л Е Г К О |

**TABLE III: EXTRACTION OF URDU CHARACTER SET BASED ON MSER AND CONNECTED COMPONENT METHOD**

| Test case number | Input | Output of text detection | Output of character segmentation |
|---|---|---|---|
| T11 | 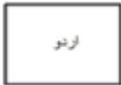 | 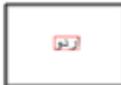 | 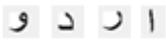 |
| T12 | 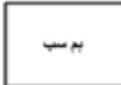 | 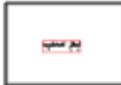 | 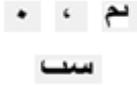 |
| T13 | 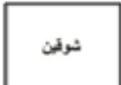 | 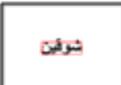 | 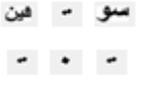 |
| T14 | 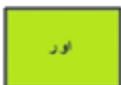 | 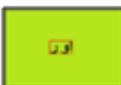 | 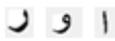 |
| T15 | 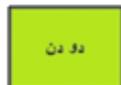 | 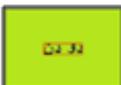 | 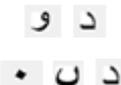 |

**TABLE IV: EXTRACTION OF HINDI CHARACTER SET BASED ON MSER AND CONNECTED COMPONENT METHOD**

| Test case number | Input | Output of text detection | Output of character segmentation |
|---|---|---|---|
| T16 | 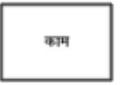 | 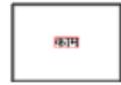 | 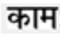 |
| T17 | 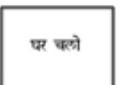 | 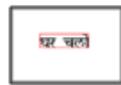 | 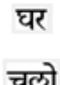 |
| T18 | 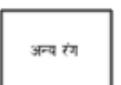 | 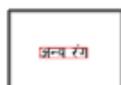 | 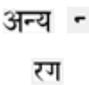 |
| T19 | 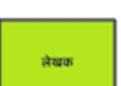 | 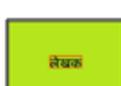 | 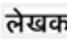 |
| T20 | 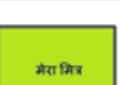 | 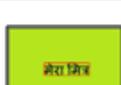 | 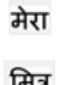 |

## V. CONCLUSION

In this paper, the primary work of identifying the textual regions using MSER followed by segmentation of each character using connected component method has been done. The proposed algorithm is tested along a set of JPEG, PNG and BMP images over four different character sets: English, Russian, Hindi and Urdu. These character sets are selected for the research as they vary in shapes, formation and writing style.

The results show that the proposed algorithm gives good results for English and Russian character set with and without background color. For Urdu and Hindi character set the text detection algorithm gives fairly good results; however the result for character segmentation algorithm which operates by exploring the connected components is not accurate. For Urdu character set it identifies the disconnected components of a single character as different characters. In Hindi character set due to presence of overline ('Shirorekha') all the characters underneath it are identified as a single component. The overall work is challenging as the compound letters are connected at various places thereby making character segmentation process difficult. Separating anuswara (.) and nuqta (.) from noise is critical as both resemble the same. Character segmentation in Hindi language is difficult due to presence of overline in it which connects all the characters under it, thereby identifying a word as a single character. All these issues can be dealt in future to improve the efficiency of the proposed algorithm and to obtain high accurac1y with various features like scaling, blur, varied font styles and font color.